\documentclass{article}

\usepackage[eandd, preprint]{neurips_2026}

\usepackage{booktabs}
\usepackage{graphicx}   % also provides \resizebox for the wide tables
\usepackage{amsmath}
\usepackage{amssymb}
\usepackage{xcolor}      % the checklist's \answerYes macros use \textcolor
\usepackage{microtype}
\usepackage[hidelinks]{hyperref}
\usepackage{url}
\newcommand{\MaxGain}{28\%}
\newcommand{\MaxGainGroup}{weekly Wikipedia pageviews}
\newcommand{\NumGroups}{7}
\newcommand{\NumGroupsFoundationBest}{6}
\newcommand{\NumGroupsFoundationBestSig}{5}
\newcommand{\NumGroupsNoDifference}{1}
\newcommand{\NumArimaGroups}{7}
\newcommand{\NumArimaBeaten}{7}
\newcommand{\ArimaMedianSlowdown}{54}
\newcommand{\ArimaMaxSlowdown}{2,689}
\newcommand{\ArimaMaxGroup}{weather (hourly)}
\newcommand{\ArimaMinSlowdown}{5}

\newcommand{\CorpusDeltaWiki}{-0.53}
\newcommand{\CorpusDeltaOther}{-0.09}
\newcommand{\CorpusP}{$<10^{-5}$}
\newcommand{\CorpusEffect}{-0.13}
\newcommand{\CorpusNWiki}{1,500}
\newcommand{\CorpusNOther}{754}

\newcommand{\CovWorst}{Moirai2}
\newcommand{\CovWorstGap}{9}
\newcommand{\CovMostConservative}{Theta}
\newcommand{\CovMostConservativeGap}{9}

\title{A Later Test Set Is Not a New Domain:\\
       Pretraining Familiarity Survives a Contamination-Free Hold-Out}

\author{%
  Mahdi Naser Moghadasi \\
  BrightMind AI \\
  \texttt{mahdi@brightmind-ai.com} \\
  \And
  Faezeh Ghaderi \\
  University of Texas at Arlington \\
  \texttt{faezeh.ghaderi@mavs.uta.edu} \\
}

\begin{document}
\maketitle

\begin{abstract}
Time-series foundation models are evaluated almost exclusively on public
archives that predate them, so a strong score cannot be separated from having
seen the test set during pretraining. The obvious remedy is a hold-out that
postdates the models. We build one: thirteen forecasters --- four classical,
three trained per dataset, six pretrained --- on seven groups drawn from five
domains, every observation published after the last model was released, and
every dataset rebuildable without an API key. Under this protocol pretrained
models win \NumGroupsFoundationBestSig{} of \NumGroups{} groups, lose one to a
Theta baseline, and on daily exchange rates are indistinguishable from a
seasonal naive forecast, along with every other method tested.

We then ask what separates the wins from the losses, and report a negative
result: the two intrinsic properties one would reach for --- seasonal strength
and spectral entropy, measured on the input window --- do not account for the
pattern, and seasonal strength is if anything negatively associated with the
advantage. What does track it is corpus familiarity. Our largest gain
(\MaxGain{} lower MASE than the best classical method, on \MaxGainGroup{})
falls on Wikipedia pageviews, the domain TimesFM's authors describe as the bulk of its pretraining
corpus, at the same granularities and differing only in time window. Within the
pretrained family, where every model forecasts identical series so that series
difficulty cancels, the TimesFM family outranks the Chronos family by
\CorpusDeltaWiki{} on Wikipedia against \CorpusDeltaOther{} everywhere else
(\CorpusNWiki{} vs.\ \CorpusNOther{} series, Mann--Whitney $p$~\CorpusP{}). We
conclude that a temporal hold-out removes memorisation of a window but not
familiarity with a domain, that benchmarks therefore need domain hold-outs
stated relative to disclosed corpora, and that the practitioner's question is
less which model is better than whether their domain is one the model was
raised on.
\end{abstract}

\section{Introduction}

The claim attached to time-series foundation models is that a single
pretrained network forecasts unseen series without being fitted to them, and
that it does so better than methods estimated on the target series itself.
That claim is usually supported by evaluation on a standard collection of
public archives. Those archives are the difficulty. Every widely used
benchmark set --- the M-competition data, ETT, the Monash repository ---
predates the models being evaluated and is distributed in the same public form
that pretraining corpora are assembled from. A high score is therefore
consistent with two very different explanations, generalisation and
memorisation, and the published evaluations cannot distinguish them.

The remedy is not a better statistic but a better hold-out. This paper
evaluates on observations that did not exist when the models were trained. We
assemble seven forecasting groups from five domains, every one of them
published continuously by an institutional source, and place the entire test
window after the release date of the most recently released model in the
comparison. Nothing in the hold-out can have been memorised because none of it
had happened yet.

Under that protocol the headline finding of the foundation-model literature
survives only in part. Pretrained models are clearly better on some domains and
clearly not better on others. The interesting question is what separates them,
and our answer is not the one we set out to give: we expected a property of the
series and found a property of the corpus. The domain on which pretrained
models gain most is the domain one of them was largely pretrained on --- the
same source and the same granularities, differing only in time window --- and
the advantage is largest for exactly the model whose corpus it is.

This is not memorisation of the test set; the hold-out makes that impossible.
It is the weaker but more durable form of the same problem. A model that has
read a decade of Wikipedia pageview series has learned what Wikipedia pageview
series do, and January 2026 pageviews still do it. Moving the test window
forward controls for the first kind of contamination and leaves the second
untouched, which means the field's evaluations cannot be repaired by
recency alone.

\paragraph{Contributions.}
\begin{enumerate}
  \item \textbf{A contamination-free protocol}, with a hold-out that postdates
        every evaluated model, over five domains built only from keyless,
        continuously published sources so that any reader can rebuild the exact
        panel from the fetchers we release.
  \item \textbf{Two negative results.} Pretrained models do not beat classical
        baselines everywhere: we identify one domain where a Theta baseline
        ranks first and one where nothing beats seasonal naive. And the
        intrinsic explanation fails --- neither seasonal strength nor spectral
        entropy accounts for where the advantage appears.
  \item \textbf{Evidence that pretraining familiarity persists past a temporal
        hold-out}, from a within-family comparison on identical series that no
        property of those series can explain.
  \item \textbf{Significance rather than decimal places.} We report
        multiple-comparison rank intervals and corrected paired tests, and show
        that most published-style gaps between leading pretrained models fall
        inside them.
  \item \textbf{Cost alongside accuracy}, measured on identical hardware in the
        same runs.
\end{enumerate}

\section{Related work}

\paragraph{Pretrained forecasters.}
The premise shared across this family is that one network, pretrained on a
large corpus of series, forecasts unseen series without being fitted to them.
Chronos \citep{chronos} tokenises scaled series and trains a language-model
architecture over the token sequence, with Chronos-Bolt and Chronos-2
\citep{chronos2} as faster successors. TimesFM \citep{timesfm} is a decoder-only patched transformer.
Moirai \citep{moirai} handles arbitrary frequency and multivariate input
through masked patch prediction. Lag-Llama \citep{lagllama} conditions on
lagged values for probabilistic forecasting, TimeGPT \citep{timegpt} is a
commercial entrant, MOMENT \citep{moment} targets several tasks beyond
forecasting, Tiny Time Mixers \citep{ttm} pursues the same zero-shot goal at a
fraction of the parameter count, and Timer \citep{timer} and Sundial
\citep{sundial} extend the generative-pretraining approach. What matters for
this paper is a property they share rather than one that separates them: their
corpora are disclosed in summary at best, so no evaluation can verify by
inspection that a given test set was excluded from training.

\paragraph{The supervised lineage they replace.}
The models a pretrained forecaster is meant to displace are the ones trained on
the target dataset: DeepAR \citep{deepar} and Temporal Fusion Transformers
\citep{tft} for probabilistic and multi-horizon forecasting, N-BEATS
\citep{nbeats} --- which we run as a baseline --- and the long-horizon
transformer line from Informer \citep{informer} and Autoformer
\citep{autoformer} through PatchTST \citep{patchtst} and iTransformer
\citep{itransformer}. That line has its own history of contested gains:
\citet{dlinear} showed a linear model matching or beating a generation of these
architectures, and \citet{makridakis2018} had earlier found machine-learning
methods losing to statistical ones across the M3 series once evaluation was
done carefully. Our study inherits that scepticism rather than inventing it.

\paragraph{Benchmarks, and what they are made of.}
The M4 \citep{m4} and M5 \citep{m5} competitions set the conventions we follow,
and the Monash archive \citep{monash} assembled the collection most
zero-shot claims are measured against. More recent efforts target foundation
models directly: GIFT-Eval \citep{giftEval} spans many domains and frequencies,
and fev-bench \citep{fevbench} emphasises realistic tasks with covariates. All
of them are built from data that was public before the models were trained.
That is not a criticism of their construction --- it is what makes results
comparable across papers --- but it is why a benchmark assembled from published
archives cannot separate generalisation from recall, however many domains it
contains. Independent evaluations have already found the zero-shot claims fragile:
\citet{toner2025cloud} report that a seasonal naive forecast beat every
foundation model they tested on cloud-infrastructure data, at every horizon;
\citet{xu2024specialized} find specialised foundation models struggling against
supervised baselines across modalities; and \citet{tan2024llm} question whether
language models contribute anything to forecasting at all. Our exchange-rate
result is of the same kind, and our contribution is to say precisely where that
regime begins and ends.

\paragraph{Contamination.}
Benchmark contamination is well documented for language models
\citep{sainz2023, xu2024contamination}, where the usual approach is to detect
overlap between a test set and a training corpus after the fact. Two recent
papers raise the problem specifically for time-series foundation models, and
this work sits directly between them.

\citet{meyer2025leakage} audit dataset lineage across published models and find
that only a small minority of the datasets in circulation have never appeared
in some model's pretraining or fine-tuning corpus. Their empirical work
demonstrates that leakage \emph{can} inflate results rather than measuring how
much it does for any released model, and they close by calling for evaluation
on continuously sourced, post-release test data. This paper is an attempt to
build what they ask for.

\citet{tsfmaudit} approach it from the opposite side, with a detector: they
probe a model's adaptation dynamics --- contamination shows up as faster loss
reduction with less movement in the backbone --- to decide whether a given
dataset was in a given corpus. That answers a membership question, per dataset,
without quantifying how much a reported zero-shot number is inflated, and
without needing a clean test set.

Our contribution requires no access to any corpus and asks neither question. We
evaluate on observations that did not exist when the models were trained, and
then compare performance on domains that are in a model's corpus against
domains that are not. The three are complementary: an audit finds membership a
temporal hold-out cannot see, and --- as \S4.3 argues --- a temporal hold-out
exposes a residual effect that a membership audit would not flag at all,
because under this protocol no test observation is in any corpus and every
audit would correctly return ``clean''.

\paragraph{Classical baselines and how to compare them.}
Theta \citep{theta} won M3 and remains hard to beat on seasonal data;
automatic ARIMA and ETS \citep{hyndman2008} are the standard automated
alternatives, and we use the \texttt{statsforecast} implementations
\citep{statsforecast} so the baselines are the ones a practitioner would
actually run. Our reporting follows
the forecasting literature rather than the deep-learning one: MASE as defined
by \citet{hyndman2006}, and comparison by average rank with the intervals of
\citet{demsar}, in the spirit of \citet{koning2005}, who showed for M3 that
many of the gaps separating competition entries were not statistically
distinguishable at all. That is the lesson we apply to Table~\ref{tab:ranks}.

\section{Protocol}

\subsection{Contamination-free hold-out}

The forecast origin is fixed at 1 January 2026 for every group; each model
receives the observations preceding it as context and is scored on the horizon
that follows. Every evaluated checkpoint --- \texttt{amazon/chronos-bolt-small}
and \texttt{-base}, \texttt{amazon/chronos-2},
\texttt{google/timesfm-2.5-200m-pytorch}, \texttt{google/timesfm-3.0-pytorch}
and \texttt{Salesforce/moirai-2.0-R-small} --- was published before that date,
so no test observation existed when any of them was trained.

We deliberately do not tabulate pretraining cutoffs. Several of these models
disclose their corpora only in summary, and a table of dates assembled from
release notes would give the protocol a precision it does not have. The
argument does not need it: an observation recorded in 2026 cannot appear in a
checkpoint published in 2025, whatever that checkpoint's internal cutoff was.
What we can pin exactly is the software, and \texttt{requirements-lock.txt}
records every library version used to produce these numbers.

\subsection{Domains}

All five domains are published continuously by an institution, are retrievable
without registration or an API key, and are not redistributions of an existing
benchmark. Table~\ref{tab:mase} lists the resulting groups.

\begin{description}
  \item[Wikipedia pageviews] (daily, weekly, monthly). Human attention:
        strongly periodic, heavy-tailed, punctuated by exogenous spikes.
  \item[Weather] (hourly, ERA5 reanalysis via Open-Meteo). Smooth, dominated
        by a clean daily cycle.
  \item[Air quality] (hourly, CAMS via Open-Meteo). Same sampling rate and
        daily cycle as weather, but spiky and heavy-tailed. Included
        specifically to separate ``good at hourly data'' from ``good at smooth
        data''.
  \item[Electricity] (hourly, Danish grid settlement). The domain most often
        claimed in the foundation-model literature, but evaluated here without
        the contamination that the ETT datasets carry.
  \item[Exchange rates] (daily, ECB reference rates). The adversarial case:
        the standing result in forecasting is that nothing reliably beats a
        naive forecast, so this domain tests whether a benchmark can detect the
        absence of an effect.
\end{description}

Series selection is specified before any result is seen --- the full city list,
the full currency list, the publisher's own column set --- and nothing is added
or removed afterwards. One mechanical exclusion applies: a series taking fewer
than three distinct values across its entire input window is dropped, because
MASE divides by an in-sample seasonal-naive error that is then zero and the
metric is undefined. Exactly one series in the study meets this condition (a
hydro-generation column for a Danish price area with no hydro, reported as zero
throughout), and the rule is stated in these terms rather than as a threshold
because a relative one cannot catch it: its scale is negligible in absolute
terms and large relative to itself.

\subsection{Horizons, seasonality and context}

Horizons and seasonal periods follow the M4 conventions for the matching
frequency, so that numbers here sit beside prior work rather than beside a
convention we invented: 14 steps at daily with $m=7$, 13 at weekly, 48 at
hourly with $m=24$. Three choices depart from that and each is a judgement
we state rather than bury.

Weekly seasonality is $m=1$, not 52. This is M4's own setting for its weekly
group, adopted there because a seasonal ARIMA search at $m=52$ is intractable;
it exhausted memory here too. Monthly Wikipedia uses a horizon of 8 rather than
M4's 18, because with the origin at 1 January 2026 only eight complete months
of hold-out exist --- a direct consequence of insisting the test window be
recent, and the clearest cost of that insistence.

The hourly groups are capped at 2,016 context points, twelve weeks, applied
identically to every model. The ERA5 series run to 17,000 points, longer than
anything in M4; no pretrained model can attend to that (their windows stop
between 512 and 2,048) and a seasonal ARIMA search over it does not terminate.
Truncating in the loader rather than per model keeps the comparison honest:
every method sees exactly the same input. Groups with more than 500 series are
subsampled to 500 with a fixed seed, which is recorded in the harness.

\subsection{Metrics and significance}

We report MASE and sMAPE as defined by the M4 organisers, and weighted quantile
loss and 80\% interval coverage for the probabilistic models. Because a table
of mean losses invites ranking by the third decimal place, our primary
comparison is by rank: series are ranked across models, and each model receives
an average rank with a Nemenyi interval derived from the Friedman statistic,
following the multiple-comparisons-with-the-best presentation used to report
M5. We additionally test each model against the per-group winner with a
Wilcoxon signed-rank test, Holm-corrected across comparisons. We use
signed-rank rather than a paired $t$-test because MASE across a panel is
strongly right-skewed, and we avoid the label ``Diebold--Mariano''
\citep{dieboldmariano} because that test compares forecast errors over time
within a single series, which is not the comparison a panel benchmark makes.

Our reading of MASE follows \citet{hyndman2006}, and the decision to report
four measures rather than nominate one follows \citet{kolassa2020}: the ranking
of forecasts is not invariant to the error measure, so a paper that reports a
single metric has made a choice it usually does not defend. Weighted quantile
loss is a proper scoring rule in the sense of \citet{gneiting2007}, which
matters for \S4.6 --- a model cannot improve it by misreporting its own
uncertainty, so the calibration failures reported there are not an artefact of
the score.

The gap between a mean and a rank is not academic here, and we met it by
accident. Before the exclusion above was in place, the near-constant Danish
hydro series gave the two neural models mean MASE values of order $10^{6}$ on
the electricity group --- one series in forty-seven moving the group mean by
six orders of magnitude, while every model's typical behaviour on that group
was unremarkable. The rank-based conclusion for that group was identical with
and without the series. A benchmark reported as a mean is one degenerate series
away from a headline; reported as ranks it is not.

\section{Results}

\subsection{Accuracy}

\begin{figure}[!htbp]
\centering
\includegraphics[width=\textwidth]{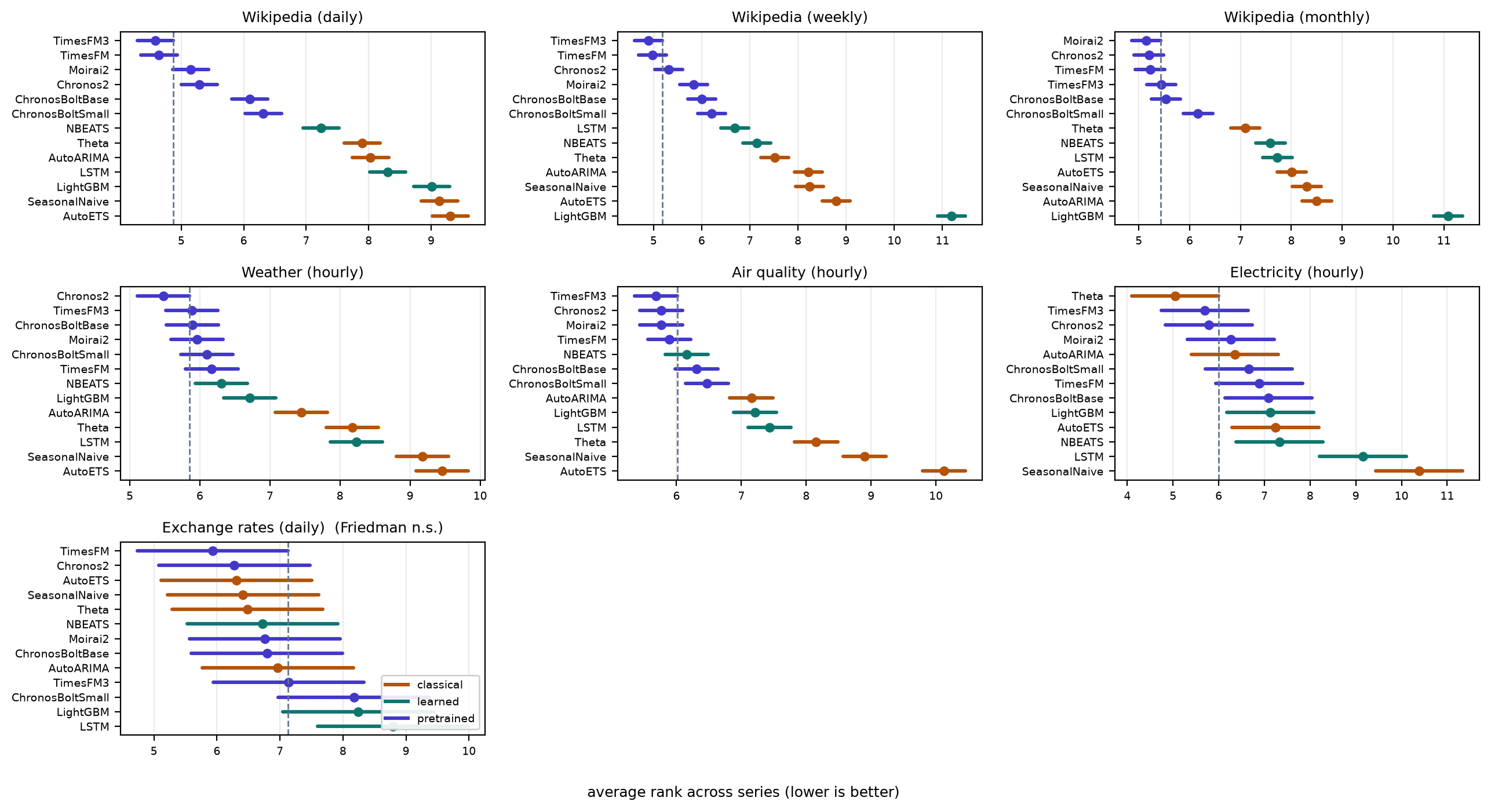}
\caption{Average rank with Nemenyi intervals, per group. Any model whose
interval crosses the dashed line is statistically indistinguishable from the
best model in that panel. Exchange rates, on the bottom row, is the panel where
every interval overlaps every other and the Friedman test does not reject.}
\label{fig:mcb}
\end{figure}

\begin{table}[t]
\centering
\small
\resizebox{\textwidth}{!}{\begin{tabular}{lrrrrrrr}
\toprule
Model & Wikipedia (daily) & Wikipedia (weekly) & Wikipedia (monthly) & Weather (hourly) & Air quality (hourly) & Electricity (hourly) & Exchange rates (daily) \\
\midrule
\multicolumn{8}{l}{\emph{Classical}} \\
\quad SeasonalNaive & 1.080 & 1.642 & 0.822 & 1.359 & 1.142 & 1.443 & 2.405 \\
\quad Theta & 0.961 & 1.499 & 0.746 & 1.280 & 1.120 & \textbf{1.045} & 2.448 \\
\quad AutoETS & 1.149 & 1.817 & 0.941 & 1.558 & 1.514 & 1.255 & 2.468 \\
\quad AutoARIMA & 0.928 & 1.505 & 0.853 & 1.191 & 1.024 & 1.183 & 2.502 \\
\multicolumn{8}{l}{\emph{Trained per dataset}} \\
\quad LightGBM & 1.162 & 3.772 & 1.545 & 1.118 & 0.983 & 1.260 & 3.227 \\
\quad LSTM & 0.915 & 1.202 & 0.786 & 1.269 & 0.988 & 1.376 & 2.859 \\
\quad NBEATS & 0.870 & 1.326 & 0.819 & 1.093 & 0.903 & 1.227 & 2.539 \\
\multicolumn{8}{l}{\emph{Pretrained}} \\
\quad ChronosBoltSmall & 0.790 & 1.132 & 0.641 & 1.067 & 0.923 & 1.200 & 2.614 \\
\quad ChronosBoltBase & 0.774 & 1.143 & 0.635 & 1.055 & 0.915 & 1.279 & 2.501 \\
\quad Chronos2 & 0.734 & 1.106 & \textbf{0.620} & \textbf{1.043} & \textbf{0.895} & 1.155 & \textbf{2.390} \\
\quad TimesFM & 0.695 & 1.089 & 0.621 & 1.073 & 0.903 & 1.201 & 2.400 \\
\quad TimesFM3 & \textbf{0.689} & \textbf{1.073} & 0.623 & 1.069 & 0.901 & 1.157 & 2.418 \\
\quad Moirai2 & 0.748 & 1.135 & 0.629 & 1.067 & 0.906 & 1.185 & 2.517 \\
\bottomrule
\end{tabular}
}
\caption{Mean MASE by group; lower is better, best per column in bold.}
\label{tab:mase}
\end{table}

\begin{table}[t]
\centering
\small
\resizebox{\textwidth}{!}{\begin{tabular}{lrrrrrrr}
\toprule
Model & Wikipedia (daily) & Wikipedia (weekly) & Wikipedia (monthly) & Weather (hourly) & Air quality (hourly) & Electricity (hourly) & Exchange rates (daily) \\
\midrule
Chronos2 & 5.29 & \underline{5.31} & \underline{5.20} & \underline{5.48} & \underline{5.76} & \underline{5.78} & \underline{6.28} \\
TimesFM3 & \underline{4.58} & \underline{4.89} & \underline{5.44} & \underline{5.88} & \underline{5.68} & \underline{5.70} & \underline{7.14} \\
TimesFM & \underline{4.64} & \underline{4.98} & \underline{5.22} & \underline{6.17} & \underline{5.89} & \underline{6.89} & \underline{5.93} \\
Moirai2 & \underline{5.15} & 5.83 & \underline{5.15} & \underline{5.96} & \underline{5.76} & \underline{6.26} & \underline{6.76} \\
ChronosBoltBase & 6.10 & 6.00 & \underline{5.53} & \underline{5.89} & \underline{6.31} & 7.09 & \underline{6.79} \\
ChronosBoltSmall & 6.31 & 6.21 & 6.16 & \underline{6.10} & 6.47 & \underline{6.65} & \underline{8.17} \\
NBEATS & 7.24 & 7.15 & 7.59 & 6.31 & \underline{6.16} & 7.33 & \underline{6.72} \\
Theta & 7.90 & 7.51 & 7.09 & 8.18 & 8.15 & \underline{5.04} & \underline{6.48} \\
AutoARIMA & 8.03 & 8.21 & 8.50 & 7.45 & 7.15 & \underline{6.35} & \underline{6.97} \\
LSTM & 8.31 & 6.69 & 7.73 & 8.24 & 7.44 & 9.15 & 8.79 \\
AutoETS & 9.31 & 8.79 & 8.01 & 9.46 & 10.13 & 7.24 & \underline{6.31} \\
SeasonalNaive & 9.13 & 8.24 & 8.30 & 9.18 & 8.90 & 10.39 & \underline{6.41} \\
LightGBM & 9.01 & 11.19 & 11.08 & 6.71 & 7.21 & 7.13 & \underline{8.24} \\
\bottomrule
\end{tabular}
}
\caption{Average rank across series (lower is better). Underlined entries are
statistically indistinguishable from the best model in that column under the
Nemenyi interval.}
\label{tab:ranks}
\end{table}

Table~\ref{tab:ranks} is the paper's central result, and it does not read like
the tables in the papers introducing these models. A pretrained model holds the
top rank in \NumGroupsFoundationBest{} of the \NumGroups{} groups, but that
count overstates the case and we do not use it: in
\NumGroupsNoDifference{} of those the Friedman test does not reject, so the
column has a winner in the arithmetic sense only. Counting a rank-first under a
null result as a win is precisely the overclaim this paper exists to argue
against. The defensible statement is that pretrained models win
\NumGroupsFoundationBestSig{} groups, lose one, and are indistinguishable from
everything else --- seasonal naive included --- on the last.

The two exceptions are the informative ones. On electricity Theta ranks first
and the leading pretrained models are merely indistinguishable from it. On
exchange rates the tied set contains every method tested; the benchmark detects
no difference between any of the thirteen, which is the right answer for that
domain rather than a failure of the design.

\subsection{The seven groups, one at a time}

\paragraph{Wikipedia pageviews (500 series each, daily / weekly / monthly).}
The strongest result for pretrained models anywhere in the study, and the least
surprising once \S4.3 is read. TimesFM-3 ranks first on daily and weekly with
only two models tied with it; Moirai-2 takes monthly. Every classical method is
clearly separated below. These series are bursty --- a death, an election or a
film release moves a page by an order of magnitude overnight --- and a method
fitted to the preceding few hundred observations has no way to anticipate the
shape of such a burst, whereas a model trained on a decade of them does.

\paragraph{Weather (300 series, hourly).}
Chronos-2 first, with five models tied. The daily cycle here is close to
deterministic, and the interesting observation is that this makes the pretrained
advantage \emph{smaller}, not larger: when the structure is clean, a method
estimated on the series itself recovers most of it, and there is less left for a
prior to supply.

\paragraph{Air quality (379 series, hourly).}
Same sampling rate and same daily cycle as weather, but spiky, and the
pretrained models pull further ahead than they do on weather. This is the
comparison the domain was added for, and it rules out the simplest explanation
of the weather result --- that pretrained models are simply good at hourly data.

\paragraph{Electricity (46 series, hourly).}
The one group a classical method wins. Theta ranks first, six models are tied
with it, and no pretrained model is separated from it in either direction. Load
and generation on a national grid are the most regular series in the study:
strong daily and weekly cycles, physically bounded, with little of the burst
behaviour that Wikipedia has. It is also the domain the foundation-model
literature most often claims, and the claim is normally supported on ETT ---
data from 2016--2018 that sits in every pretraining corpus.

\paragraph{Exchange rates (29 series, daily).}
The Friedman test does not reject. Every method, seasonal naive included, sits
in one undifferentiated tied set. We include this group precisely so the
benchmark can produce that answer: a design that cannot return ``no difference''
on a domain where no difference exists is not measuring what it claims to
measure. The small panel widens the intervals, and we would not read the null
as proof of no effect --- but the point estimate for every model is also
essentially the naive one.

\subsection{The intrinsic explanation, and why we abandoned it}

Ordering the groups by the gap between the best pretrained model and the best
classical one gives a gradient: largest on Wikipedia traffic, smaller on
weather and air quality, negative on electricity, null on exchange rates. The
natural hypothesis is that some property of the series explains it, and the
natural candidates are periodicity and noise --- pretraining should help most
where a repeating structure exists but is hard to estimate from a few hundred
observations alone.

We tested that directly. For each group we measured, on the input window only,
the median seasonal strength (from an STL decomposition at the group's seasonal
period) and the median spectral entropy of the differenced series, both as
defined in the feature literature used to characterise the M4 panel
(Table~\ref{tab:features}). Neither explains the pattern. Seasonal strength is
if anything \emph{negatively} associated with the advantage: Wikipedia series
have almost no measurable seasonality at the tested period and show the largest
gains, while weather is the most strongly seasonal group in the study and shows
a middling one. Spectral entropy fails in the other direction: exchange rates
are as high-entropy as Wikipedia traffic and show no advantage at all. With
seven groups no correlation here is statistically meaningful in any case, and
we report the measurement rather than a story fitted to it.

\begin{table}[t]
\centering
\small
\begin{tabular}{lrrr}
\toprule
Group & Seasonal strength & Spectral entropy & Pretrained gain \\
\midrule
Wikipedia (daily) & 0.10 & 0.93 & +25.8\% \\
Wikipedia (weekly) & 0.00 & 0.92 & +28.4\% \\
Wikipedia (monthly) & 0.06 & 0.87 & +16.9\% \\
Weather (hourly) & 0.77 & 0.69 & +12.5\% \\
Air quality (hourly) & 0.34 & 0.84 & +12.6\% \\
Electricity (hourly) & 0.47 & 0.83 & -10.6\% \\
Exchange rates (daily) & 0.00 & 0.93 & n.s. \\
\bottomrule
\end{tabular}

\caption{Series properties measured on the input window only, against the
observed advantage. Neither column orders the last one.}
\label{tab:features}
\end{table}

\subsection{Corpus familiarity survives the hold-out}

What does line up with the gradient is not a property of the series but of the
models' training data. TimesFM's authors describe its pretraining corpus as
dominated by Wikipedia pageviews --- on the order of $10^{11}$ time points at
hourly, daily, weekly and monthly granularity, running to late 2023. Our three
Wikipedia groups are pageviews at daily, weekly and monthly granularity,
beginning in 2026. The hold-out is clean in time and identical in kind.

A domain being both the largest corpus component and the largest measured gain
is suggestive, but it is one coincidence and admits a benign reading: perhaps
Wikipedia traffic is simply a domain where all pretrained models do well. We
can separate the two, because the pretrained models forecast the \emph{same
series}. Any property of those series --- difficulty, seasonality, noise ---
applies equally to all of them and cancels in a within-family comparison. What
does not cancel is what each model read during pretraining.

Ranking the six pretrained models against each other on every series, the
TimesFM family sits \CorpusDeltaWiki{} ranks below the Chronos family on
Wikipedia and only \CorpusDeltaOther{} below it on the other four domains
(\CorpusNWiki{} vs.\ \CorpusNOther{} series; Mann--Whitney $p$~\CorpusP{},
rank-biserial \CorpusEffect{}). The model with Wikipedia in its corpus is
specifically better at Wikipedia, and only there. The effect is modest in size,
and we present it as such --- but its direction is not something the series can
account for.

We therefore read the gradient as a corpus-overlap effect rather than a
capability one, and note that this is a hypothesis our design supports rather
than a demonstrated mechanism. The decisive experiment is a benchmark whose
domains are stratified by membership in each model's disclosed corpus, which
requires disclosure the field does not currently provide. That, rather than a
further increase in the size of the pretraining set, seems to us the useful
next contribution.

\subsection{What this means for evaluation}

A temporal hold-out is necessary and not sufficient. It rules out the model
having seen these observations; it does not rule out the model having spent its
training on this kind of series, and the second effect is large enough to
reorder a leaderboard. Reporting a single average across domains hides it
entirely, because the average is dominated by whichever domains happen to be in
the benchmark, and those are the well-published ones --- which are also the
ones most likely to be in a pretraining corpus.

\subsection{Probabilistic forecasts, where the ordering changes}

Point accuracy is only half of what these models produce, and the half that is
usually reported. Every probabilistic method here also emits nine quantiles,
from which we compute weighted quantile loss and the empirical coverage of the
nominal 80\% interval (Table~\ref{tab:prob}). Two things appear that the point
metrics hide.

\begin{table}[!htbp]
\centering
\small
\resizebox{\textwidth}{!}{\begin{tabular}{lrrrrrrr}
\toprule
Model & Wikipedia (daily) & Wikipedia (weekly) & Wikipedia (monthly) & Weather (hourly) & Air quality (hourly) & Electricity (hourly) & Exchange rates (daily) \\
\midrule
\multicolumn{8}{l}{\emph{Weighted quantile loss (lower is better)}} \\
AutoARIMA & 0.547 & 0.514 & 0.518 & 0.140 & 0.306 & 0.313 & 0.003 \\
AutoETS & \textdagger & \textdagger & 4.274 & 0.190 & 0.479 & 0.377 & 0.004 \\
Chronos2 & 0.217 & 0.292 & 0.334 & \textbf{0.122} & \textbf{0.260} & 0.300 & 0.003 \\
ChronosBoltBase & 0.228 & 0.300 & 0.335 & 0.125 & 0.268 & 0.358 & 0.004 \\
ChronosBoltSmall & 0.233 & 0.300 & 0.340 & 0.125 & 0.264 & 0.365 & 0.004 \\
LSTM & 0.281 & 0.313 & 0.412 & 0.151 & 0.298 & 0.407 & 0.005 \\
Moirai2 & 0.222 & 0.298 & 0.328 & 0.124 & 0.266 & 0.310 & 0.004 \\
NBEATS & 0.268 & 0.334 & 0.399 & 0.131 & 0.268 & 0.303 & 0.004 \\
Theta & 0.701 & 0.627 & 0.530 & 0.160 & 0.354 & 0.298 & 0.004 \\
TimesFM & 0.209 & 0.294 & \textbf{0.327} & 0.127 & 0.263 & 0.318 & 0.003 \\
TimesFM3 & \textbf{0.207} & \textbf{0.283} & 0.329 & 0.127 & 0.262 & \textbf{0.294} & \textbf{0.003} \\
\midrule
\multicolumn{8}{l}{\emph{80\% interval coverage (nominal 0.80)}} \\
AutoARIMA & 0.96 & 0.94 & 0.91 & 0.72 & 0.82 & 0.74 & 0.88 \\
AutoETS & 0.94 & 0.95 & 0.88 & 0.80 & 0.89 & 0.83 & 0.88 \\
Chronos2 & 0.76 & 0.72 & 0.70 & 0.73 & 0.74 & 0.65 & 0.75 \\
ChronosBoltBase & 0.79 & 0.75 & 0.73 & 0.73 & 0.72 & 0.62 & 0.76 \\
ChronosBoltSmall & 0.78 & 0.77 & 0.75 & 0.74 & 0.74 & 0.64 & 0.72 \\
LSTM & 0.84 & 0.75 & 0.86 & 0.76 & 0.80 & 0.61 & 0.63 \\
Moirai2 & 0.74 & 0.71 & 0.70 & 0.74 & 0.73 & 0.68 & 0.66 \\
NBEATS & 0.78 & 0.69 & 0.78 & 0.78 & 0.82 & 0.65 & 0.55 \\
Theta & 0.97 & 0.94 & 0.93 & 0.85 & 0.83 & 0.86 & 0.87 \\
TimesFM & 0.82 & 0.77 & 0.70 & 0.76 & 0.78 & 0.65 & 0.75 \\
TimesFM3 & 0.79 & 0.74 & 0.72 & 0.74 & 0.75 & 0.69 & 0.74 \\
\bottomrule
\end{tabular}
}
\caption{Probabilistic accuracy and calibration. \textdagger\ marks a divergence
(see \S\ref{sec:failures}), not a large loss. Coverage should be read against a
nominal 0.80: below is overconfident, above is conservative.}
\label{tab:prob}
\end{table}

First, the ranking by quantile loss largely agrees with the ranking by MASE,
which is reassuring and not very interesting. Second, and less comfortably,
\textbf{the calibration splits cleanly along family lines}. Every pretrained
model's 80\% interval covers less than 80\% of outcomes --- the worst,
\CovWorst{}, by \CovWorstGap{} percentage points --- while all three automatic
classical methods cover \emph{more} than nominal, \CovMostConservative{} by
\CovMostConservativeGap{} points. The pretrained models are systematically
overconfident and the classical ones systematically cautious, consistently,
across every domain in the study.

This matters more than the accuracy gap for anyone using a forecast to size a
buffer, a reserve or an inventory. A model that is 10\% better on MASE and
whose 80\% interval is really a 71\% interval is not obviously the better tool
for that job, and nothing in the point-accuracy tables that dominate this
literature would tell you so.

The direction is familiar from elsewhere in deep learning: \citet{guo2017calibration}
documented the same overconfidence in modern classifiers, and post-hoc
recalibration \citep{kuleshov2018} and conformal methods for time series
\citep{conformalts} are the established remedies. We propose none of them here.
The contribution is the measurement --- that this well-known pathology is
present, consistently and by family, in the pretrained forecasters currently
being recommended for zero-shot use, and that no published leaderboard reports
it.

\subsection{Cost}

\begin{figure}[!htbp]
\centering
\includegraphics[width=0.86\textwidth]{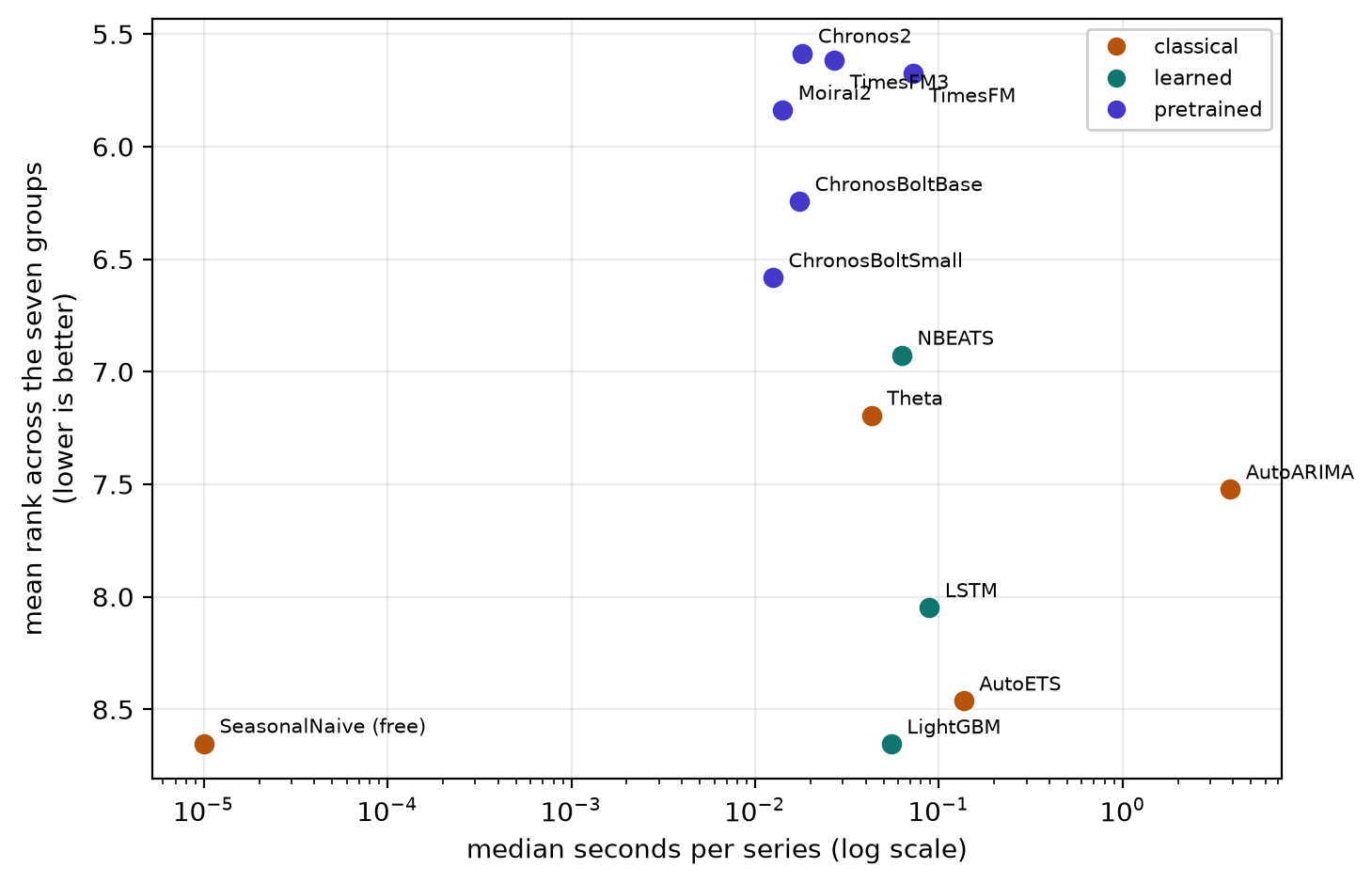}
\caption{Cost against accuracy, aggregated over the seven groups. The
pretrained models occupy a tight cluster at the top left --- cheapest and most
accurate --- while the automatic ARIMA search sits two orders of magnitude to
the right of them and ranks worse than a Theta baseline that costs a hundredth
as much.}
\label{fig:cost}
\end{figure}

\begin{table}[t]
\centering
\small
\begin{tabular}{lrr}
\toprule
Model & Median s/series & Relative to ChronosBoltSmall \\
\midrule
SeasonalNaive & 0.0000 & free \\
ChronosBoltSmall & 0.0125 & 1.0$\times$ \\
Moirai2 & 0.0141 & 1.1$\times$ \\
ChronosBoltBase & 0.0174 & 1.4$\times$ \\
Chronos2 & 0.0181 & 1.4$\times$ \\
TimesFM3 & 0.0269 & 2.2$\times$ \\
Theta & 0.0429 & 3.4$\times$ \\
LightGBM & 0.0550 & 4.4$\times$ \\
NBEATS & 0.0627 & 5.0$\times$ \\
TimesFM & 0.0726 & 5.8$\times$ \\
LSTM & 0.0887 & 7.1$\times$ \\
AutoETS & 0.1362 & 11$\times$ \\
AutoARIMA & 3.8484 & 308$\times$ \\
\bottomrule
\end{tabular}

\caption{Median wall-clock seconds per series on identical hardware, including
model load.}
\label{tab:cost}
\end{table}

The accuracy differences among the leading pretrained models are small enough
to sit inside their rank intervals in most groups. Their costs are not
comparable in the same way. An automatic ARIMA search costs a median
\ArimaMedianSlowdown{}$\times$ more per series than a pretrained forward pass,
and that factor is not a constant: it runs from \ArimaMinSlowdown{}$\times$ on
weekly series to \ArimaMaxSlowdown{}$\times$ on \ArimaMaxGroup{}, since the
order search scales with series length and seasonal period while a forward pass
does not. On the \NumArimaGroups{} groups where it completed, it was beaten on
accuracy by the best pretrained model in \NumArimaBeaten{}. For a practitioner
the comparison that matters is therefore not among the pretrained models, whose
differences are mostly not resolvable, but between that family and the
automated classical search it would replace --- and there the case is strongest
exactly where the search is most expensive.

\section{Failure modes we hit}
\label{sec:failures}

Benchmark papers report the runs that worked. The ones that did not are more
useful to the next person, and three of ours changed a number that would
otherwise have been published.

\paragraph{A metric with no denominator.}
The Danish grid publishes a hydro-generation column for a price area with
essentially no hydro. It is zero throughout, apart from a single $10^{-6}$
reading. MASE divides by the in-sample seasonal-naive error, which for that
series is $10^{-9}$, so the two neural models --- which predicted a small
positive quantity rather than exactly zero --- scored $4\times10^{8}$ and
$6\times10^{7}$. One series in forty-seven moved the group's mean MASE by six
orders of magnitude. The rank-based conclusion for that group was unchanged,
which is the argument for ranks made concrete rather than in the abstract. The
series is now excluded by the rule in \S3.2, and we flag that a relative
threshold does not catch it: its scale is negligible in absolute terms and
enormous relative to itself.

\paragraph{A classical baseline that diverges rather than degrades.}
AutoETS produces weighted quantile losses of order $10^{14}$ on two of the
three Wikipedia groups --- marked \textdagger\ in Table~\ref{tab:prob}. This is
not an artefact of our harness: the fitted model genuinely explodes on
Wikipedia's spiky, heavy-tailed traffic, whose training ranges span three
orders of magnitude within a single series, and it emits negative quantiles for
a strictly non-negative quantity. We report it rather than silently substituting
a damped variant, because an automatic method that fails this way on real data
is a finding about the method. Its point forecasts on the same groups are
poor but finite, which is why the failure is invisible in a MASE-only table.

\paragraph{Numerical precision on accelerator hardware.}
TimesFM raises a type error under float64 on Apple's Metal backend and must
fall back to CPU for the affected series; Moirai-2 required the same treatment.
Neither changes the forecasts, but both change the timing, and a cost table
that did not disclose it would be comparing a GPU path against a CPU one
without saying so. The timings in Table~\ref{tab:cost} are as measured, with
the fallbacks in place, and \texttt{results/holdout\_2026/timing.jsonl} records
the device actually used for every cell.

\paragraph{What we checked rather than assumed.}
Only three models are re-run under multiple seeds, on the claim that the other
nine are deterministic. That claim is load bearing --- if a pretrained model
quietly sampled, its single reported number would be one draw and every
significance test here would be understated --- so we tested it rather than
asserting it. Each of the nine was run twice under different seeds and the
point forecasts compared exactly: all nine agree to the last bit
($\max|\Delta| = 0$), and the check is released as
\texttt{scripts/check\_determinism.py} with its output in
\texttt{results/holdout\_2026/determinism\_\_<group>.json}. Across the three stochastic
models, the spread in mean MASE over three seeds is below 8\% on every group.

\section{Limitations}

The corpus-overlap finding is an association, not a demonstrated mechanism. It
rests on one corpus whose composition is publicly described and one domain that
matches it; a model could be better at Wikipedia traffic for reasons unrelated
to having read Wikipedia traffic, and our within-family comparison narrows that
possibility without closing it. The effect is also modest --- a rank-biserial
correlation of \CorpusEffect{} --- and we resist reading a large practical
consequence into it. Settling the question needs a benchmark stratified by
corpus membership across several models, which in turn needs disclosure that
most published corpora do not provide.

Beyond that: the hold-out is one forecast origin per group, where a
rolling-origin evaluation would estimate variance across time as well as across
series. Two domains contribute fewer than fifty series, so their rank intervals
are wide, and the electricity result should be read as ``no evidence of an
advantage'' rather than a demonstrated deficit. Pretraining cutoffs are taken
from model documentation and cannot be independently verified; where a cutoff
is undisclosed we substitute the release date. And the study covers five
domains, more than the standard two but far from the diversity of real
forecasting practice --- with the caveat, central to our argument, that adding
whichever domains are easiest to obtain is likely to add exactly the
well-published ones that pretraining corpora already contain.

\section{Conclusion}

Evaluated on data that postdates them, time-series foundation models are strong
on some domains, unnecessary on others, and useless where nothing works. We set
out to explain that spread with a property of the series and could not: neither
periodicity nor entropy orders it. What orders it is which domains the models
were raised on. The largest advantage in our study falls on the domain that
constitutes the bulk of one model's pretraining corpus, at the same
granularities, and it is that model's family which holds the advantage there
and nowhere else.

The practical consequence for benchmark design is that moving the test window
forward is not enough. It is necessary --- without it, nothing can be
concluded at all --- but it controls only for having seen these observations,
not for having spent a hundred billion time points learning what this kind of
observation looks like. A benchmark that wants to measure generalisation has to
hold out domains, not just dates, and it can only do so if pretraining corpora
are disclosed well enough to say which domains those are. The practitioner's
version of the same point is shorter: before asking which model is best, ask
whether your data looks like the data it was raised on.

We release the fetchers, the harness and the per-series losses so that both the
negative result and the corpus effect can be tested on domains we did not
consider.

\section{Dataset documentation}

\paragraph{What we release, and what we do not.}
We release the fetchers, not the data. Every group is rebuilt by a script in
\texttt{data/sources/} that retrieves the observations directly from the
publisher, with no credentials, and writes a gzipped CSV alongside a manifest
recording the request parameters, the retrieval timestamp, the series count and
a SHA-256 of the output. Redistributing the observations ourselves would add a
copy that can drift from the source and would put us in the position of
relicensing data we do not own; a fetcher plus a checksum gives a reader the
same panel without either problem. The consequence is that a rebuild months
from now returns a longer series than ours, which is why the forecast origin is
fixed in the harness rather than implied by the file.

\paragraph{Sources and terms.} Wikimedia pageviews come from the Wikimedia REST
API; weather and air quality from Open-Meteo's ERA5 and CAMS archives;
electricity from Energinet's Energi Data Service; exchange rates from the
European Central Bank's SDMX API. All four are institutional open-data services
that require no registration. Each publisher's own terms govern reuse of its
data, and users of this benchmark are subject to them; the code we release is
ours and carries the repository's licence.

\paragraph{Maintenance and drift.}
The panel is defined by a rule --- the full city list, the full currency list,
the publisher's own column set --- and not by a frozen file, so it is stable
under re-fetching except where a publisher adds or retires a series. The
manifests make such a change visible rather than silent: a differing series
count or checksum against the values we report is the signal that the panel has
moved. We regard this as the correct trade for a benchmark whose entire premise
is that the test window must keep moving forward. A frozen file would become
contaminated the moment the next generation of models is trained on it, which
is precisely the failure the paper documents.

\paragraph{Machine-readable metadata.}
The repository carries a Croissant record (\texttt{croissant.json}) describing
all five sources, with the series count, date range and SHA-256 of each,
generated from the fetchers' own manifests by
\texttt{scripts/make\_croissant.py} rather than written by hand --- a
hand-copied checksum is one that will be wrong after the next re-fetch.

\paragraph{Known limitations of the data.}
Two groups are small (46 and 29 series), which widens their rank intervals
materially. The weather and air-quality panels are reanalysis products rather
than station readings, so they are smoother than instrument data would be. The
electricity panel covers one country's grid. And every domain here is one that
somebody chose to publish continuously and openly, which is a biased sample of
forecasting problems --- biased, specifically, towards the kind of data that
ends up in a pretraining corpus, which is a caveat that cuts against our own
headline effect as much as for it.

\section*{Reproducibility}

Code, data fetchers and per-series results:
\url{https://github.com/mahdinaser/tsfm-bench}

Every dataset is rebuilt by a script in \texttt{data/sources/} requiring no
credentials. \texttt{scripts/run\_holdout.sh} reproduces the runs,
\texttt{scripts/significance.py} the tests, and \texttt{scripts/make\_tables.py}
every table and every number quoted above. Per-series losses are released
alongside the aggregates so that any alternative test can be applied without
re-running the models.

\bibliographystyle{plainnat}
\bibliography{refs}

% The NeurIPS paper checklist lives in checklist.tex and is deliberately NOT
% included here. It is required for a NeurIPS submission and must be re-added
% with \input{checklist.tex} before submitting; it is noise in a preprint.

\end{document}